\documentclass[twocolumn]{style}
\runninghead{SMILES Summer School Projects Proceedings}
\footertext{\textit{SMILES} 2025}

\usepackage[most]{tcolorbox}
\usepackage{fontawesome5}
\usepackage{xcolor}

\newtcolorbox{takeawaybox}{
  enhanced,
  breakable,
  colback=blue!10!white,        
  colframe=blue!40!white,       
  coltitle=black,               
  fonttitle={\small\bfseries},  
  title={\faAtom\hspace{0.5em}Takeaway},
  fontupper=\small,             
  boxrule=0.4pt,
  arc=3pt,
  left=8pt,
  right=8pt,
  top=6pt,
  bottom=6pt,
  before skip=10pt,
  after skip=10pt
}

\title{Token Homogenization under Positional Bias}  
\logo{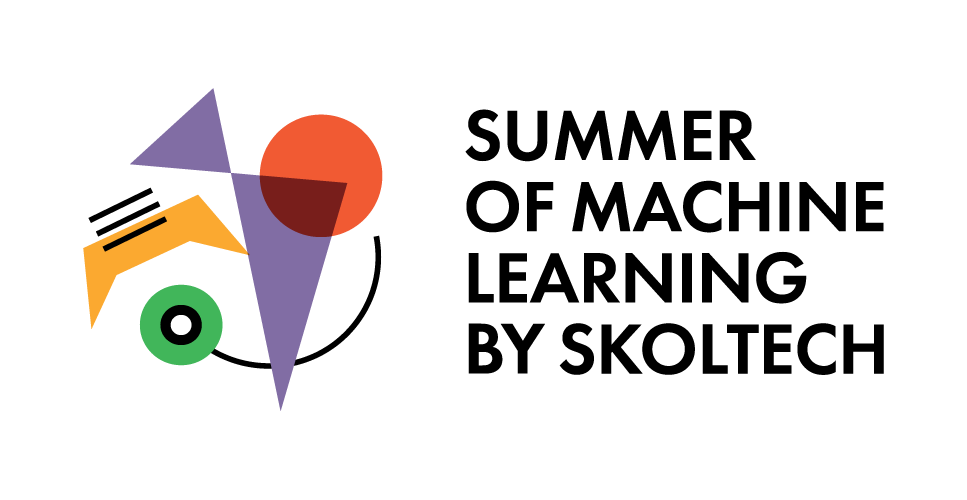}

\author{
    \textbf{Viacheslav Yusupov*}\textsuperscript{2, 3}, \textbf{Danil Maksimov*}\textsuperscript{2, 3},
    \textbf{Ameliia Alaeva*}\textsuperscript{1, 3}, 
    \textbf{Tatiana Zaitceva*}\textsuperscript{4},
    \textbf{Antipina Anna*}\textsuperscript{5}, 
    \textbf{Anna Vasileva*} \textsuperscript{3}\\
    Chenlin Liu \textsuperscript{6},
    Rayuth Chheng\textsuperscript{7},
    Danil Sazanakov\textsuperscript{8},
    Andrey Chetvergov\textsuperscript{9, 10, 11},
    \\
    \textbf{Project curators:} Egor Shvetsov\textsuperscript{12}, 
    Alina Ermilova\textsuperscript{12} \\
    \textbf{*Asterisk signifies equal contribution} \\
}

\date{
    \textsuperscript{\textbf{1}}
    International Laboratory of Bioinformatics, HSE University,
    \textsuperscript{\textbf{2}}
    Laboratory for Matrix and Tensor Computations in Machine Learning, HSE University,
    \textsuperscript{\textbf{3}}
    Faculty of Computer Science, HSE University,
    \textsuperscript{\textbf{4}}
    Department of Mechanics and Mathematics, Lomonosov Moscow State University,
    \textsuperscript{\textbf{5}}
    Faculty of Space Research, Lomonosov Moscow State University ,
    \textsuperscript{\textbf{6}}
    Harbin Institute of Technology ,
    \textsuperscript{\textbf{7}}
    Institute Of Technology Of Cambodia,
    \textsuperscript{\textbf{8}}
    Faculty of Geography and Geoinformation Technology, HSE University,
    \textsuperscript{\textbf{9}}
    ISP RAS,
    \textsuperscript{\textbf{10}}
    Artificial Intelligence Research Center, RANEPA,
    \textsuperscript{\textbf{11}}
    ITMO University 
     \textsuperscript{\textbf{12}}
     Skoltech
}

\begin{document}
\twocolumn[\maketitle]
\small
\begin{abstract}
   \noindent This paper investigates \textit{token homogenization}---the convergence of token representations toward uniformity across transformer layers---and its relationship to positional bias in large language models. We empirically examine whether homogenization occurs and how positional bias amplifies this effect. Through layer-wise similarity analysis and controlled experiments, we demonstrate that tokens systematically lose distinctiveness during processing, particularly when biased toward extremal positions. Our findings confirm both the existence of homogenization and its dependence on positional attention mechanisms.

\end{abstract}



\section{Introduction}

In this work, we demonstrate a possible link between two previously known phenomena in Large Language Models (LLMs): 1) \textit{positional bias} and 2) \textit{representation homogenization}.

\begin{figure}[h]
\centering
\includegraphics[width=1\linewidth]{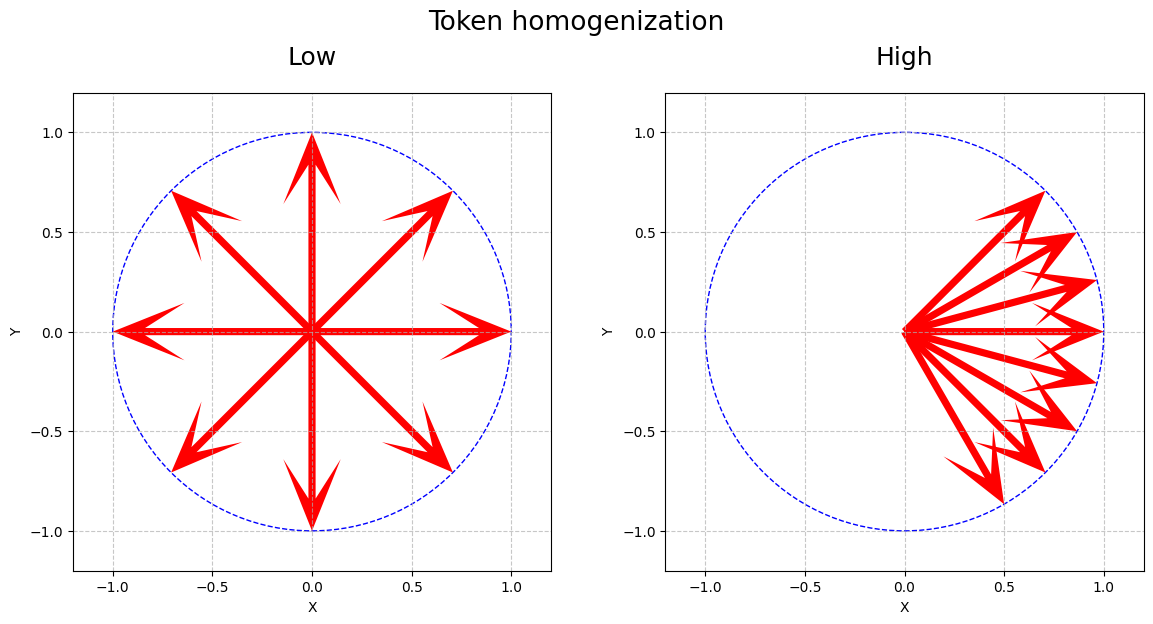}
\caption{An illustration of low and high anisotropy for vectors in $2D$ space.}
\label{fig:resultant_length}
\end{figure}

\paragraph{Representation homogenization and layer-wise dynamics.}
This paper refers to \textit{homogenization} as a process wherein repeated information mixing operations in deep neural architectures progressively diminish representational diversity\footnote{A more rigorous definition of token homogenization in transformers accounting for positional bias is presented in Section~\ref{sec:positional_bias}.}. As the definition of representational diversity depends on the applied context and admits, prior research has documented this phenomenon through distinct metrics.

A phenomenon known as rank collapse has been observed in deep self-attention-based models~\cite{dong2021attention, geshkovski2025mathematical, wu2405role, naderi2024mind}, where repeated application of attention layers may project token representations into a rank-$1$ vector space. This phenomenon extends beyond self-attention architectures, a similar representation dynamic called \textit{over-smoothing}~\cite{di2022understanding, keriven2022not} occurs in Graph Neural Networks.

\begin{takeawaybox}
The key intuition is that attention matrices "mix" information, and repeated mixing converges to a space that is uninformative—we refer to this process as \textit{homogenization}.
\end{takeawaybox}

\noindent Homogenization in transformers arises fundamentally because attention matrices progressively average token information. \citeauthor{vitvitskyi2025makes}~\cite{vitvitskyi2025makes} formally proved that when generalizing to longer contexts at inference time, global attention matrices inevitably converge to uniform "pure mixing" states. Separately, \cite{barbero2410round} showed decoder-only transformers exhibit heightened sensitivity to initial tokens due to causal masking and further documented \textit{representational collapse} where information from later tokens is systematically degraded in long sequences.

Concurrently, several works have analyzed the representation dynamics in causal architectures. \citeauthor{lee2024geometric}~\cite{lee2024geometric} reveal layer-specific compression: during training, the number of significant PCA components decreases, indicating progressive dimensionality reduction. Notably, compression rates vary across layers, with later layers in larger models exhibiting slower reduction. In the foundational work, \citeauthor{ethayarajh2019contextual}~\cite{ethayarajh2019contextual} characterized \textit{anisotropy of the representation} in transformers, defined geometrically as word representations occupying "a narrow cone in vector space rather than spreading uniformly" (illustrated in Figure~\ref{fig:anis}). To quantify these dynamics, \cite{ethayarajh2019contextual}  employ metrics such as self-similarity, intrasentence similarity, and maximum explainable variance (MEV), the latter of which we adopt.

\begin{takeawaybox}
Anisotropy intuition: Token representations occupy a narrow cone in the vector space rather than spreading uniformly in all directions.
\end{takeawaybox}

Together, these effects point towards a key difficulty: Transformers tend to "over-mix" information both as they become deeper \cite{barbero2410round} and as they ingest longer contexts \cite{vitvitskyi2025makes}.

\paragraph{Positional bias in transformer-based LLMs.}
LLMs exhibit \textit{positional bias}—systematic neglect of information in specific context positions—which manifested as U-shaped attention patterns prioritizing extremal tokens while degrading mid-context processing~\cite{zhang2024found, wu2025emergence, liu2023lost}. This bias stems from architectural factors (e.g., causal masking) and training data biases~\cite{barbero2025llms, guo2024serial}. The authors~\cite{mikhail2025position} demonstrate that positional bias degrades model performance and hypothesizes a connection to token homogenization.

In this work, we investigate how positional bias drives \textit{token homogenization}: the convergence of token representations toward uniformity across transformer layers. While formal analyses suggest this relationship~\cite{mikhail2025position}, empirical evidence is lacking. This gap raises two critical questions:
\begin{itemize}
\item \textbf{Q1}: How can we measure homogenization?
\item \textbf{Q2}: How does positional bias amplify this effect?
\end{itemize}

\begin{figure}
    \centering
    \includegraphics[width=1\linewidth]{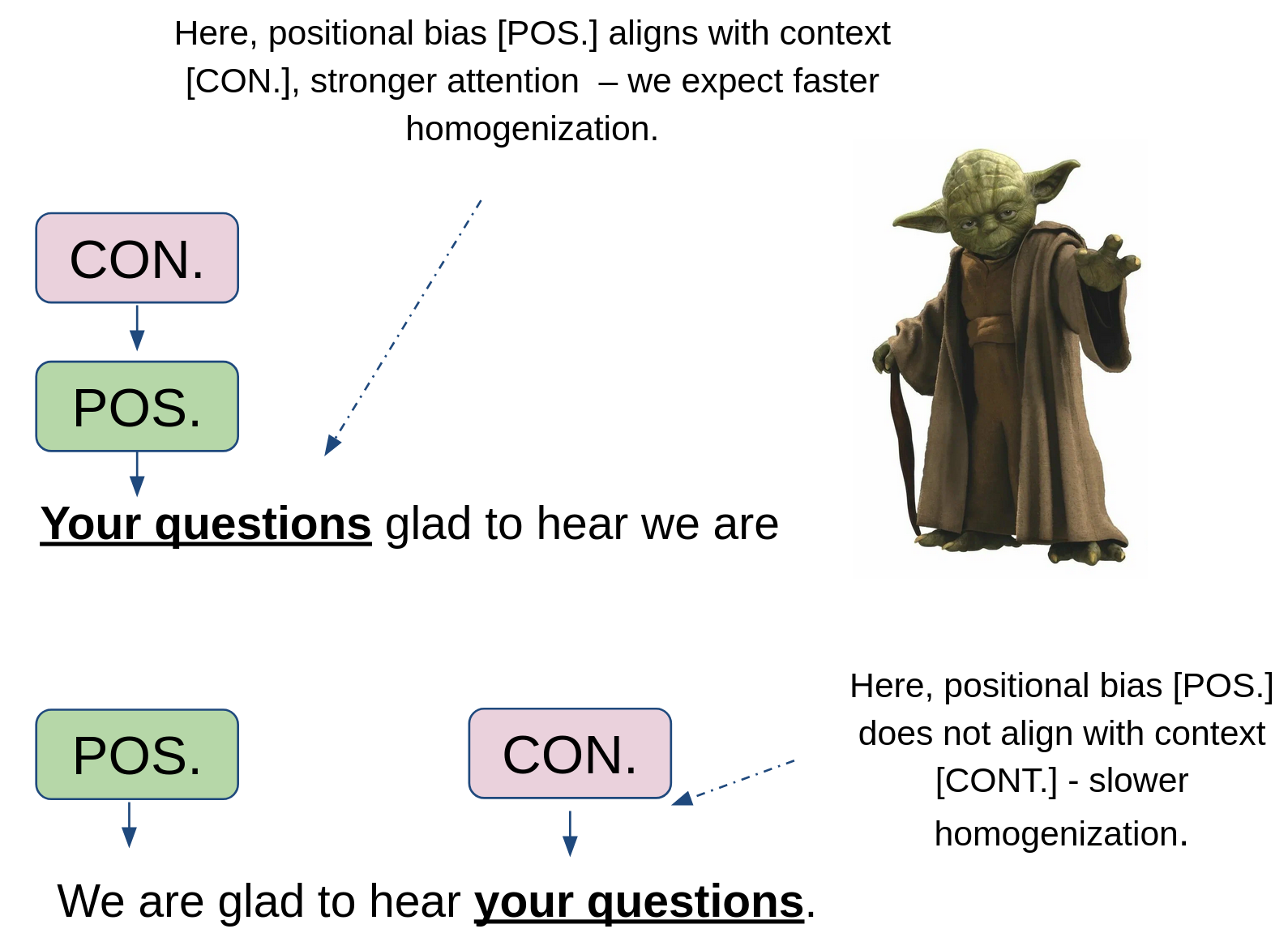}
    \caption{The illustration how positional bias amplifies homogenization. Assume positional bias is present at first tokens, therefore, if the most important word is present at the beginning of a sentence  it would increase attention weights of this word and therefore amplify homogenization. Their positions could differ. The "CON." denotes the most important part of the context -  contextual attention,  "POS." denotes the positional bias - positional attention.}
    \label{fig:bias-illustration}
\end{figure}

In Figure \ref{fig:bias-illustration} we illustrate how homogenization can be amplified with positional bias. Assume we have two additive attention types - contextual and positional, if they align the total attention value would increase, therefore, amplifying  mixing. The formal description of this effect is provided in Section~\ref{sec:positional_bias}.

\vspace{25pt}

\paragraph{Contributions:}
\begin{itemize}
    \item In this paper we demonstrate a new set of metrics to analyse model's activations, this set of metrics can be efficiently employed for mechanistic interpretability, for example, we are the first to analyse LLMs activations with MAUVE scores. 
    \item Using a new set of metrics, we demonstrate a connection between positional bias and token homogenization for three models, LLaMA-3, Gemma and Qwen. 
\end{itemize}

\section{Token Homogenization under Positional Bias}
\label{homogenization}

\label{sec:positional_bias}
Here, we formalize how positional bias in transformer-based LLMs drives token homogenization across layers similarly to \cite{mikhail2025position}. Our idea of how positional bias is connected to contextual attention is represented in Figure~\ref{fig:bias-illustration}. Assuming a standard multi-head self-attention architecture:

\textbf{Notation:} Let \( X^{(0)} = [x_1, x_2, \dots, x_n] \in \mathbb{R}^{n \times d} \) denote the input token embeddings, where \( x_1 \) is the first token and \( d \) is the embedding dimension. In layer \( l \geq 1 \), self-attention computes:
\[
A^{(l)} = \text{softmax}\left(X^{(l)} W_Q^{(l)} (X^{(l)} W_K^{(l)})^\top\right), 
\]
\[
X^{(l+1)} = W_O^{(l)} A^{(l)} X^{(l)} W_V^{(l)},
\]
where \( W_Q^{(l)}, W_K^{(l)}, W_V^{(l)} \in \mathbb{R}^{d \times d}, W_O^{(l)} \in \mathbb{R}^{n \times n} \) are projection matrices.

\textbf{Assumptions:} To isolate homogenization effects:  
(1) Positional attention \(A^{\mathrm{pos}}\) is concentrated on the first token:
\[
A_{i,j}^{\mathrm{pos}(l)} \approx 
\begin{cases} 
1 & \text{if } j = 1 \\  
0 & \text{otherwise}
\end{cases}
\]
(2) Positional bias persists across layers and  
(3) Attention decomposes as \( A^{(l)} = \lambda_1 A^{\mathrm{cont}(l)} + \lambda_2 A^{\mathrm{pos}(l)} \) with \( \lambda_1 + \lambda_2 = 1 \).

\textbf{Homogenization Mechanism:} Token representations evolve as:
\[
x_i^{(l)} \approx \lambda_2 P^{(l)} x_i^{(l-1)} + \lambda_1 x_i^{\mathrm{con}(l)}, \quad P^{(l)} = W_O^{(l)} W_V^{(l)}
\]
When \( \lambda_2 > \lambda_1 \), recursive application across \( l \) layers yields:
\[
x_i^{(l)} \approx x_1^{(l)} \quad \forall i
\]
\vspace{0.2cm}
\noindent\textit{Interpretation:} This demonstrates explicit homogenization, where all tokens converge to the initial first-token embedding. While residual connections prevent complete collapse in practice, positional bias systematically increases token similarity through layer-wise propagation.
\section{Methods}
\label{methods}
Positional bias is often observed at the beginning or end of prompts in large language models (LLMs)\cite{mikhail2025position}. To explore the potential connections between positional bias and homogenization, we aim to determine whether homogenization increases when key words are placed at the beginning or the end of prompts. This is illustrated in Figure\ref{fig:bias-illustration}. To achieve this, we restructured a movie review dataset to systematically position the most informative words at either the beginning (labeled as "Front") or the end (labeled as "End") of each prompt. From this point forward, we refer to these modified datasets as \textbf{Front} and \textbf{End}, respectively. The detailed procedure for the dataset creation is described in Section~\ref{sec:data}.

The next step involves quantifying the degree of homogenization itself. To accomplish this, we introduce a suite of metrics that includes both novel measures and established methodologies; all metrics will be detailed in Section~\ref{sec:metr}.

\subsection{Models} We analysed modern LLM models of different families such as the LLaMa-3 8B \cite{openlm2023openllama, touvron2023llamaopenefficientfoundation},  Gemma 7B \cite{team2024gemma} and Qwen-2.5 7B \cite{qwen2.5} models.

\subsection{Dataset creation}

\label{sec:data}

For the original dataset, we took a subsample of size $1000$ movie reviews from IMDB \cite{imdb2011}.  
First, we utilized the GLM-4-Flash-250414 model to identify the most important words in movie reviews. Then we used Qwen-3-235B \cite{yang2025qwen3technicalreport} to rephrase the reviews so the most important part is placed at the \textbf{Front} or in the \textbf{End}. Examples of samples from synthetic datasets, as well as prompts for paraphrasing, could be found in Section \ref{appendix:dataset_generation}.

\begin{table} 
\begin{center}
\begin{tabular}{|c||c|} 
   \hline
   Metric  & BERTScore (F1) \\
   \hline
   Front dataset  & 0.9008  \\
   \hline
   End dataset  & 0.8964  \\
   \hline
\end{tabular}
\end{center}
\caption{Metrics of semantic similarity, calculated between dataset with key words at the front of the paragraph and original dataset (2nd line) and between dataset with key words at the end of the paragraph and original dataset (3rd line). }
\label{table:datasets_similarity}
\end{table}

To verify that rephrasing paragraphs has not altered their meaning, we calculated the mean BERTScore \cite{zhang2019bertscore} for $2$ paraphrased datasets and an original one and provided the results of the comparison in Table \ref{table:datasets_similarity}. This metric of semantic similarity was selected because its value is not affected by words permutation, since the text organization in original and paraphrased paragraphs may differ substantially. The proposed metric takes values from $0$ to $1$, and higher values of the metric indicate a closer semantic similarity. 

To evaluate the spelling and syntactic correctness, as well as the overall naturalness of the generated texts, we calculated their perplexity and compared it against that of the original dataset with the LLM models we applied. The results of this comparison are provided in Table \ref{table:datasets_similarity}. We have also noticed that synthetic datasets tend to have substantially higher mean and maximum per-token entropy than an original one, indicating that higher token homogenization implies per-token entropy growth. 

\begin{table}
\begin{center} 
\begin{tabular}{|c||c|c|c|}
   \hline
   Model  & LLaMa-3 8B & Gemma 7B & Qwen-2.5 7B \\
   \hline
   Original dataset  & 15.91 & 15.99 & 16.18 \\
   \hline
   Front dataset  & 17.80 & 18.43 & 17.81 \\
   \hline
   End dataset  & 19.2 & 19.98 & 19.61 \\
   \hline
\end{tabular}
\end{center}
\caption{Perplexities of the synthezised and original datasets, calculated on different LLM models}
\label{table:datasets_quality}
\end{table}

\subsection{Metrics}
\label{sec:metr}
As discussed in the introduction, homogenization admits varying definitions depending on the application context. This section introduces a set of metrics that we employ to identify conditions under which the connection between homogenization and positional bias becomes the most pronounced.

\textbf{Singular Values} \label{singval} The common technique to analyse the matrices and attention matrices is to consider the singular value decomposition (SVD) and to analyse the singular values of the matrix. For matrix $A \in \mathbb{R}^{n \times m}$:
\[
A = U\Sigma V^T,
\]
where $U \in \mathbb{R}^{n \times k}$, $V \in \mathbb{R}^{m \times k}$ and $\Sigma \in \mathbb{R}^{k \times k}$ -- is the diagonal matrix with singular values $\sigma_1 \ge  \sigma_2 \ge ... \ge \sigma_k \ge 0$ on the diagonal. 

In our method, we compute the SVD for the average output token matrix $X_s^{(l)}$ of the layer $l$ of the model for a fixed data sample $s$. For the matrices $X_s^{(l)}$ we compute singular values $\sigma^{(s, l)}_i$, $i \in [1, 2, 3]$. 
Therefore, if there is a significant difference in the singular values across layers, we can infer a lack of homogenization among the layers. 
The singular values $\sigma_i$ across the layers are demonstrated in Section \ref{res}. 

\textbf{Limitation of SVD-based approaches} Metrics, derived from the SVD of internal representations matrix share a disadvantage: they falsely positive claim token homogeneity if tokens have almost opposite direction. This issue can be practically important for the antonyms, occurring in the same sentence, as the representations of the antonymous tokens tend to have oppositely directed representations \cite{lingua_repr}. For instance, imagine that we have a sentence consisting of 2 words, each corresponding to a token: \textit{"Good bad"}. The internal representation after layer $l$ of the word \textit{"good"} is $v$, of the word \textit{"bad"} is $-v$, then the representation matrix $H_{(l)} \in \mathbb{R}^{2 \times d}$ is equal to 

\[
H_{(l)} = \begin{bmatrix}
    v \\
    -v
\end{bmatrix}
\]

It would have rank-$1$, what will indicate token homogenization which is not the case.

\textbf{Maximum Explainable Variance}, essentially representing the non-uniformity of a distribution in space If word representations from a particular layer were isotropic (i.e., directionally uniform), then the average cosine similarity between uniformly randomly sampled words would be $0$ \cite{ethayarajh2019contextual}. In contrast, the closer the token representation to each other, the closer the cosine similarity between them, as well as anisotropy to $1$. Similarly to the work \cite{razzhigaev2024shapelearninganisotropyintrinsic}, we compute the Maximum Explainable Variance metric for each output token matrix $X_s^{(l)}$ of the model for sample $s$ and layer $l$:
\[
MEV(X_s^{(l)}) = \frac{\sigma_1^2}{\sum_{i = 1}^n \sigma_i^2}.
\]
With the growth of the number of layers, if homogenization occurs, the anisotropy decreases, which we preliminary observe in Section \ref{res}. 

\textbf{Schatten norms} To aggregate the values of singular values, we consider the Schatten norms of the matrices $X_s^{(l)}$. The Schatten $p$-norm $S_p(x)$ is defined as $l_p$-norm of the vector $x$, so in our case $S_p^{(s, l)} = \|(\sigma^{(s, l)}_1, \sigma^{(s, l)}_2, \ldots, \sigma^{(s, l)}_k)\|_p$. In case of "ideal" homogenization, the matrices $X_s^{(l)}$ have rank one, and all eigenvalues except the largest one should be zero. So, we could approximate the eigenvalues as $(\lambda, \varepsilon, \ldots, \varepsilon)$. 
In this case $S_1 \approx \lambda + (k-1) \varepsilon$, $S_2 \approx \sqrt{\lambda^2 + (k-1) \cdot \varepsilon^2} \approx \lambda + \lambda(k-1)\varepsilon^2/2$, $S_{\infty} = \lambda$. Therefore, in case of homogenization, we expect that $S_1$ should be larger than $S_2$ and $S_\infty$. 

An intuitive approach to the quantify token homogenization effect involves computing the \textit{rank} of the matrix $X_s^{(l)} \in \mathbb{R}^{n \times d}$ formed by the $n$ token representations in the layer $l$. A rank deficiency ($\text{rank}(X_s^{(l)}) \ll d$) would suggest the representation collapse. However, this method suffers from two critical limitations:

\begin{enumerate}
    \item \textbf{Numerical Instability}:  
    Rank computation via singular value decomposition (SVD) is sensitive to numerical thresholds. For $X_s^{(l)}$ with rapidly decaying but non-zero singular values $\{\sigma_i\}$, slight perturbations (e.g., floating-point errors or noise) can drastically alter rank estimates, making this metric unreliable.
    
    \item \textbf{Magnitude Ignorance}: 
    Standard rank disregards the \textit{relative significance} of different directions in the representation space. Treats all non-zero singular values equally, ignoring their magnitudes. A matrix with one dominant singular value and several near-zero values (indicating strong homogenization) may have the same rank as one with uniform singular values (indicating diverse representations).
\end{enumerate}

To address these issues, we propose using \textit{effective-rank} -- continuous, spectrum-aware metric. 

\textbf{Effective rank} \cite{roy2007effective}. \textbf{Definition}
For a matrix $X \in \mathbb{R}^{m \times n}$ with singular values $\sigma_1 \geq \sigma_2 \geq \cdots \geq \sigma_Q \geq 0$ ($Q = \min\{m,n\}$), define the normalized singular value distribution:
\[
p_k = \frac{\sigma_k}{\|\boldsymbol{\sigma}\|_1}, \quad \|\boldsymbol{\sigma}\|_1 = \sum_{k=1}^Q \sigma_k
\]
The \textbf{effective rank} is given by:
\[
\text{erank}(X) = \exp\left(H(\mathbf{p})\right) \quad \text{where} \quad H(\mathbf{p}) = -\sum_{k=1}^Q p_k \log p_k
\]
with $0 \log 0 \equiv 0$. This corresponds to the exponential of the spectral entropy.

Effective rank has 3 useful properties:
\begin{enumerate}
\item \textbf{Effective rank is a lower bound of the rank}: For $X \in \mathbb{R}^{m \times n},X \neq 0$ $\text{erank}(X) \leq \text{rk}(X)$
\item \textbf{Effective rank is invariant under scaling}: For $X \in \mathbb{R}^{m \times n}, c\neq 0$ $\text{erank}(X) = \text{rk}(cX)$, which means that scaling internal representations does not affect the effective rank metric.
\item \textbf{Effective rank is unitary invariant}. This property, in particular, indicates that our metric does not depend on the order of the internal representations of the tokens in the representation matrix $X_s^{(l)}$.
\end{enumerate}

\begin{figure*}[h]
\begin{center}
\begin{minipage}[h]{0.33\linewidth}
\center{\includegraphics[width=1\linewidth]{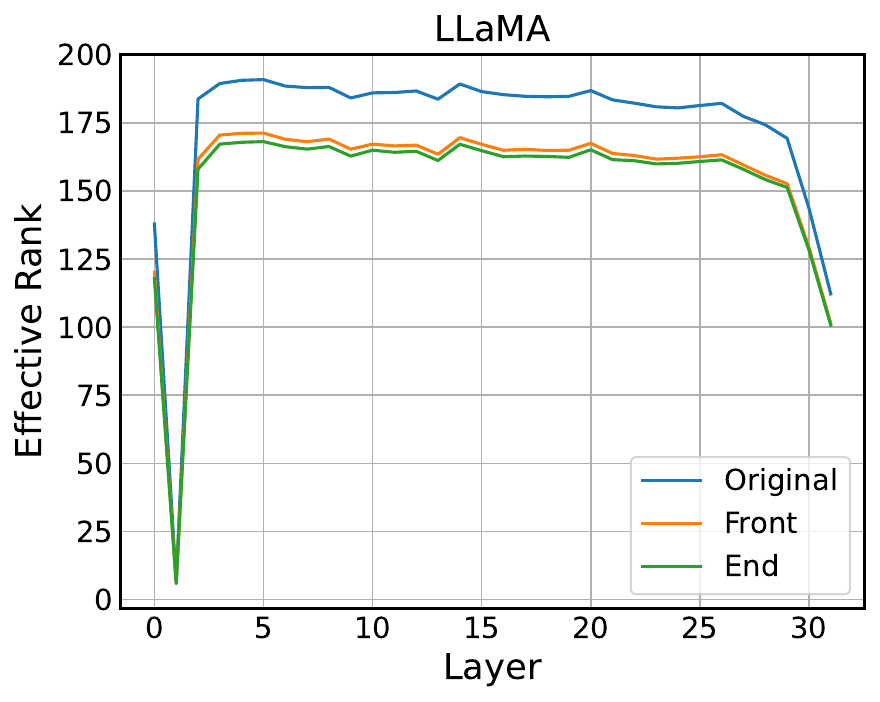}} 
\end{minipage}
\hfill
\begin{minipage}[h]{0.33\linewidth}
\center{\includegraphics[width=1\linewidth]{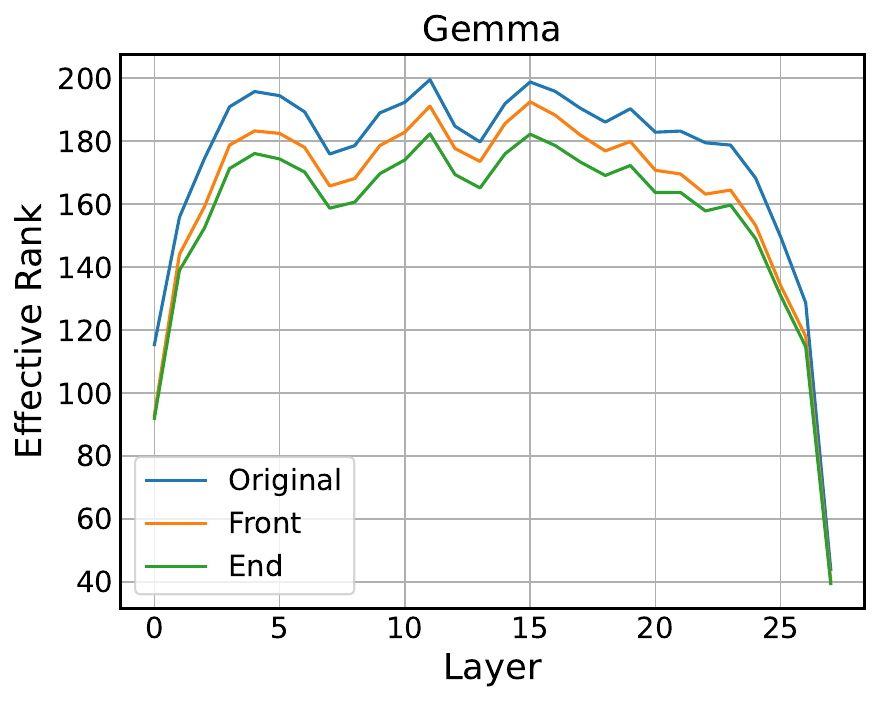}} 
\end{minipage}
\hfill
\begin{minipage}[h]{0.33\linewidth}
\center{\includegraphics[width=1\linewidth]{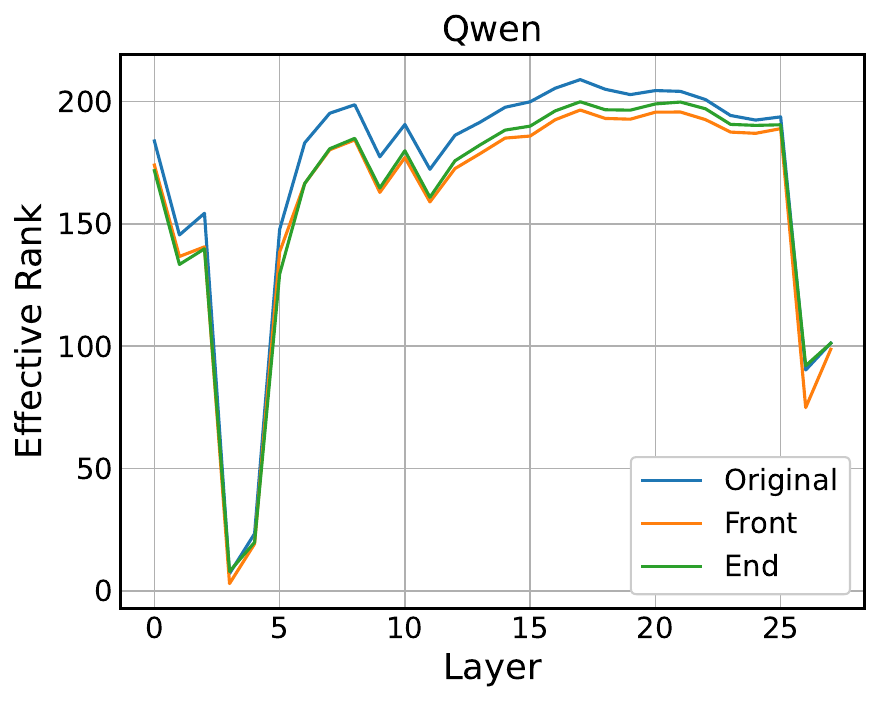}} 
\end{minipage}
\caption{The average effective rank for matrices $X_s^{(l)}$ across the layers of the LLaMa \cite{openlm2023openllama} (left), Gemma \cite{gemmateam2024gemmaopenmodelsbased} (center) and Qwen \cite{qwen2.5} (right) models on the original and synthetic datasets. The blue color denotes the original dataset, orange -- the dataset with front positional bias and green -- with end positional bias \ref{sec:data}.}
\label{fig:rank}

\end{center}
\end{figure*}

\textbf{Distribution-based Approach} 
However, there is one more approach to explore the token homogenization problem, which is based on the probability distribution. 

The values of matrix $X_s^{(l)}$  for each layer $l$ (see Singular Values~\ref{singval}) have their own probability distribution. We propose to observe the homogenization process by measuring the divergence of distributions between each layer of matrices using \textit{MAUVE score}.

\textbf{MAUVE score.} Assume we have two probability distributions $P$ and $Q$. The metric is based on the calculation of the Kullback–Leibler (KL) divergency between  $P$ and $Q$ but also includes the weighted mixture of these distributions to form the \textit{divergence curve}. The purpose of the curve is to formalize and balance information about the trade-off between Type I and II errors. \textbf{MAUVE ($P$, $Q$)} is the area under the \textit{divergence curve}. The explicit definition is provided in the following papers\cite{pillutla-etal:mauve:jmlr2023} \cite{pillutla-etal:mauve:neurips2021} \cite{liu-etal:mauve-theory:neurips2021}.

Summarizing all above, the MAUVE score measures the statistical gap between two distributions.  MAUVE ($P$, $Q$) lies in $(0, 1]$. The closer the MAUVE score to $1.0$, the more similar are $P$ and $Q$. 

Originally, this metric was proposed for text distributions, e.g., to compare texts written by a model and a human. It was found to correlate the strongest with human evaluations over baseline metrics for open-ended text generation. However, MAUVE can work with other modalities such as images, speech, music, or video, as long as domain-appropriate embeddings are obtained. There is an official implementation of this metric, which is obtained by computing the KL-divergences between the two distributions in a quantized embedding space of a foundation model.

In our problem, we assume that token embedding matrices are distributed in a uniform way among the first layers of neural network, which agrees with the non-homogenized case. When homogenization occurs, a large number of values are close or similar, which corresponds to dense distributions on the deep layers. We expect a higher MAUVE score comparing distributions of initial layers and the opposite situation for the last ones. 

\textbf{Resultant length} Another distribution-based approach for measuring token representation homogeneity is to calculate the \textit{resultant length}.

\textbf{Definition} Suppose that we have $n$ unit length $p$-dimensional vectors $x_1, ... x_n \in \mathcal{S}^{p-1}$. Then its \textbf{resultant length} $\bar{R}$ is defined as: 
\[
\bar{R} =\| \frac{x_1 + \dots + x_n}{n} \|_2, \text{where } \forall i \in \{1...n\}:  \|x_i\|_2=1 
\]

The idea of such a measure is depicted on the figure \ref{fig:resultant_length}: The more vectors that concentrate by their mean direction, the higher the resultant length. The statistical intuition behind this metric is provided in Appendix \ref{vmf}.

In our case, we normalize the lengths of internal representations of different tokens and calculate the resultant length for it as a measure of the direction similarity.

\begin{figure*}[h]
\begin{center}
\begin{minipage}[h]{0.33\linewidth}
\center{\includegraphics[width=1\linewidth]{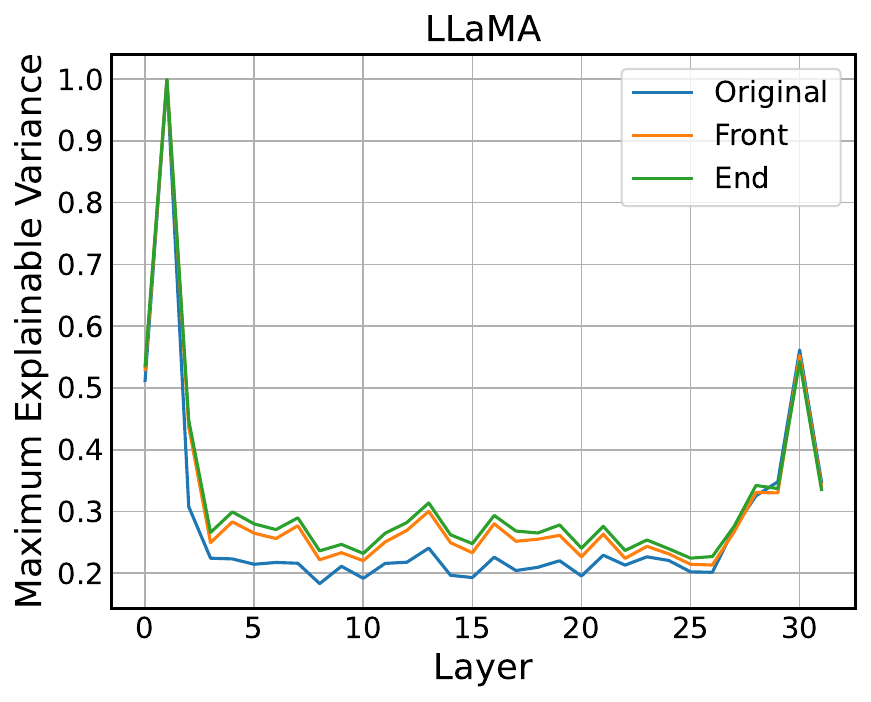}} 
\end{minipage}
\hfill
\begin{minipage}[h]{0.33\linewidth}
\center{\includegraphics[width=1\linewidth]{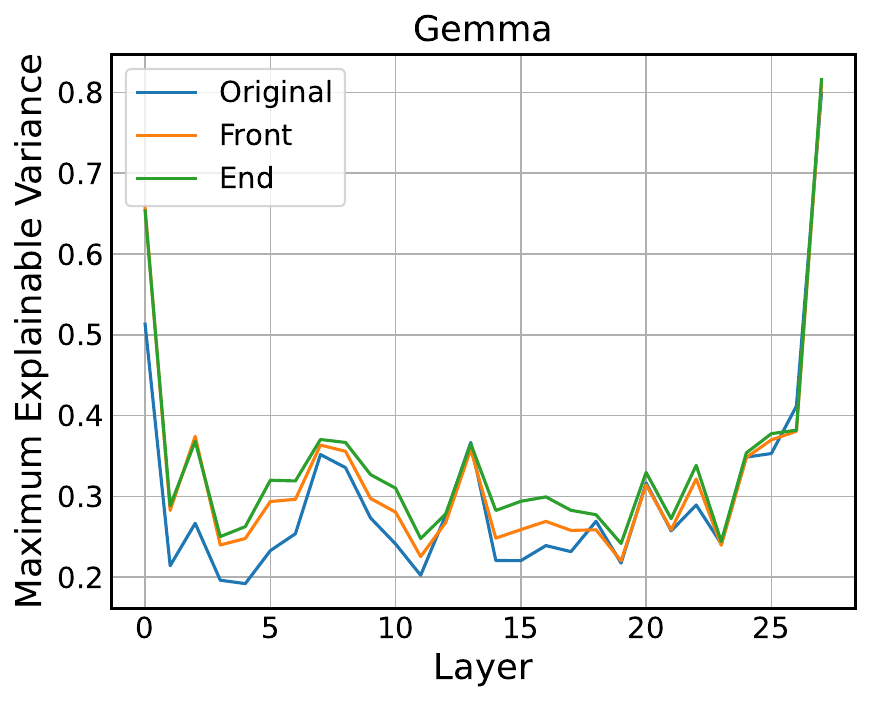}} 
\end{minipage}
\hfill
\begin{minipage}[h]{0.33\linewidth}
\center{\includegraphics[width=1\linewidth]{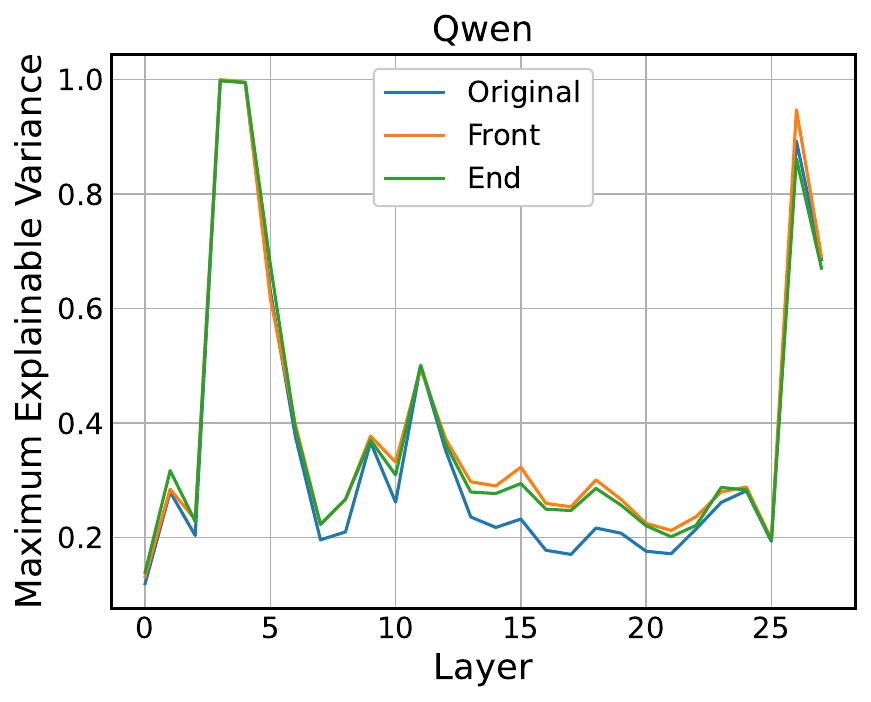}} 
\end{minipage}
\caption{The average Maximum Explainable Variance for matrices $X_s^{(l)}$ across the layers of the LLaMa \cite{openlm2023openllama} (left),  Gemma \cite{gemmateam2024gemmaopenmodelsbased} (center) and Qwen \cite{qwen2.5} (right) models on the original and synthetic datasets. The blue color denotes the original dataset, orange -- the dataset with front positional bias and green -- with end positional bias \ref{sec:data}.}
\label{fig:anis}

\end{center}
\end{figure*}

\begin{figure*}[h]
\begin{center}
\begin{minipage}[h]{0.33\linewidth}
\center{\includegraphics[width=1\linewidth]{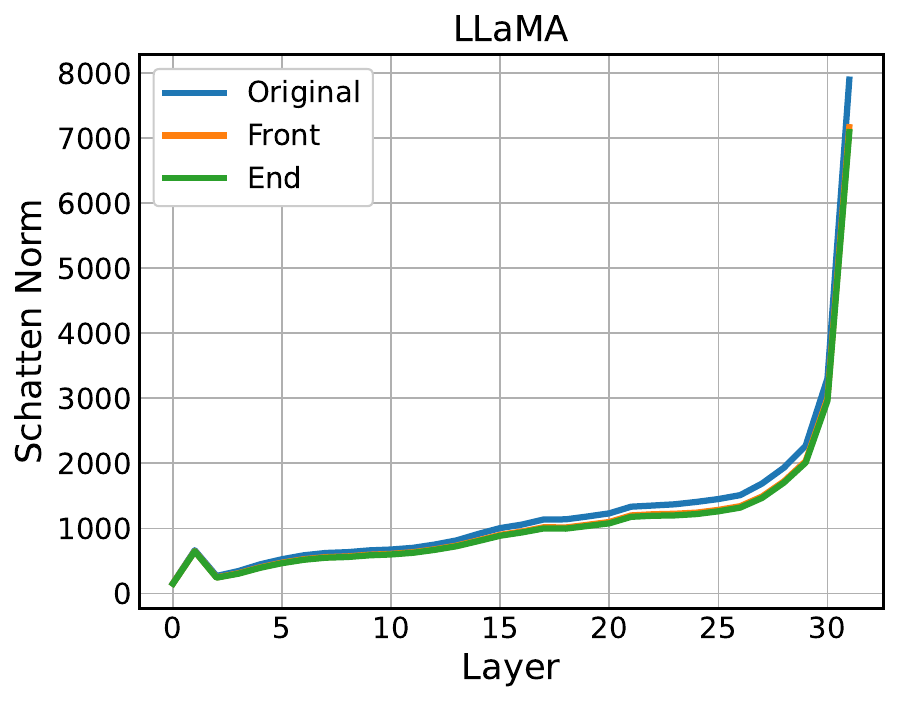}} 
\end{minipage}
\hfill
\begin{minipage}[h]{0.33\linewidth}
\center{\includegraphics[width=1\linewidth]{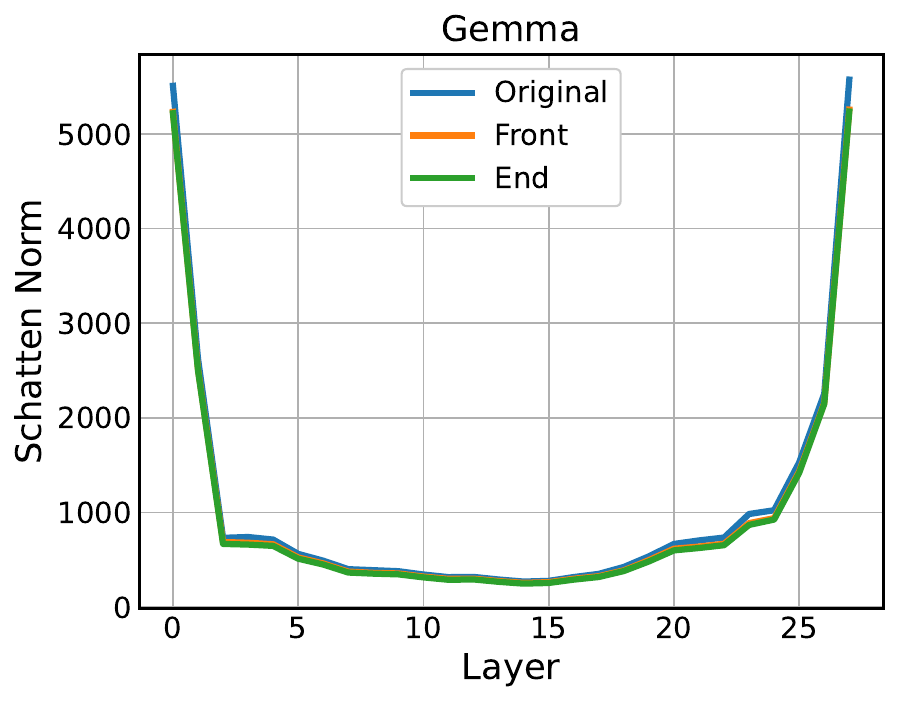}} 
\end{minipage}
\hfill
\begin{minipage}[h]{0.33\linewidth}
\center{\includegraphics[width=1\linewidth]{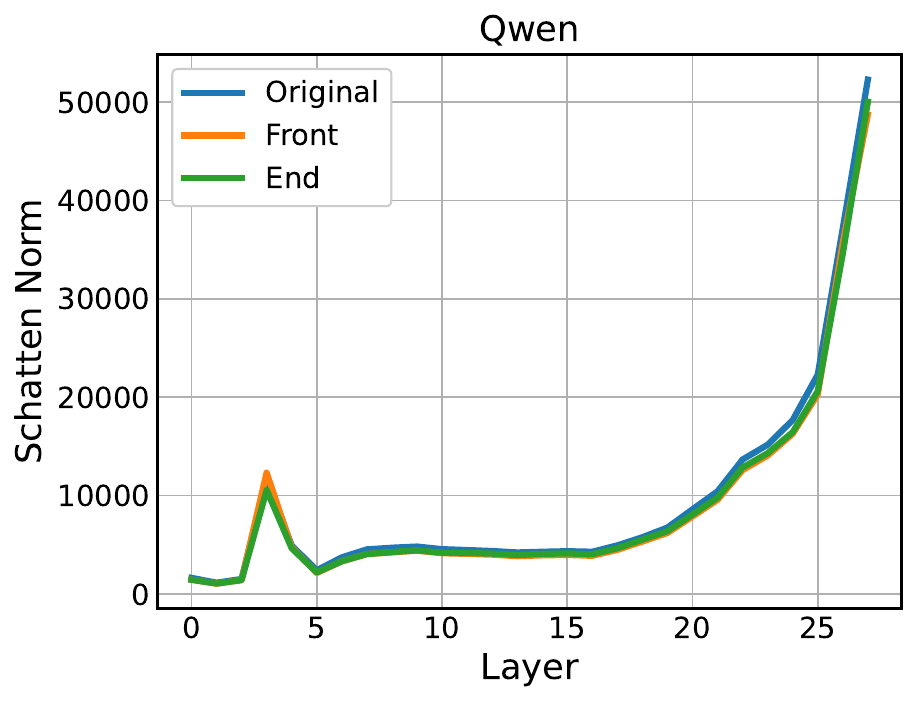}} 
\end{minipage}
\caption{The average Schatten norms for matrices $X_s^{(l)}$ across the layers of the LLaMa \cite{openlm2023openllama} (left),  Gemma \cite{gemmateam2024gemmaopenmodelsbased} (centre) and Qwen \cite{qwen2.5} (right) models on the original and synthetic datasets. The blue colour denotes the original dataset, orange -- the dataset with front positional bias and green -- with end positional bias \ref{sec:data}.}
\label{fig:shatten}

\end{center}
\end{figure*}

\subsection{Positional Bias Presence Detection}
To demonstrate the positional bias, we construct the attention maps and compute the sum of the attention values in the columns of $A_s^{(l)}$ for the sample $s$ and the layer $l$. We observed the positional bias at the end of the token sequence. The visualization of this effect is demonstrated in Section \ref{res}.


\begin{figure*}[h]
\begin{center}
\begin{minipage}[h]{0.33\linewidth}
\center{\includegraphics[width=1\linewidth]{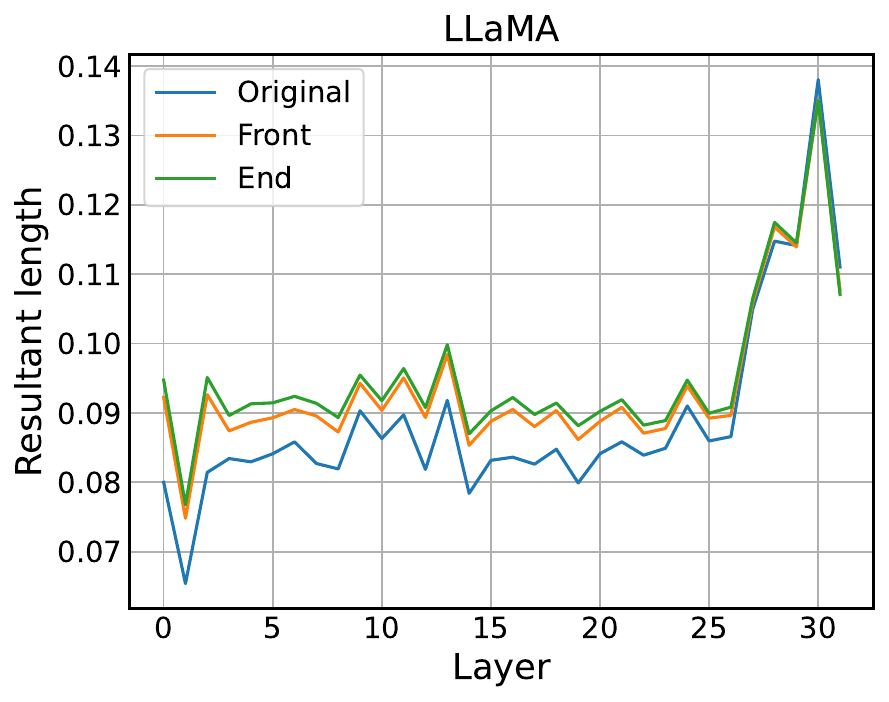}} 
\end{minipage}
\hfill
\begin{minipage}[h]{0.33\linewidth}
\center{\includegraphics[width=1\linewidth]{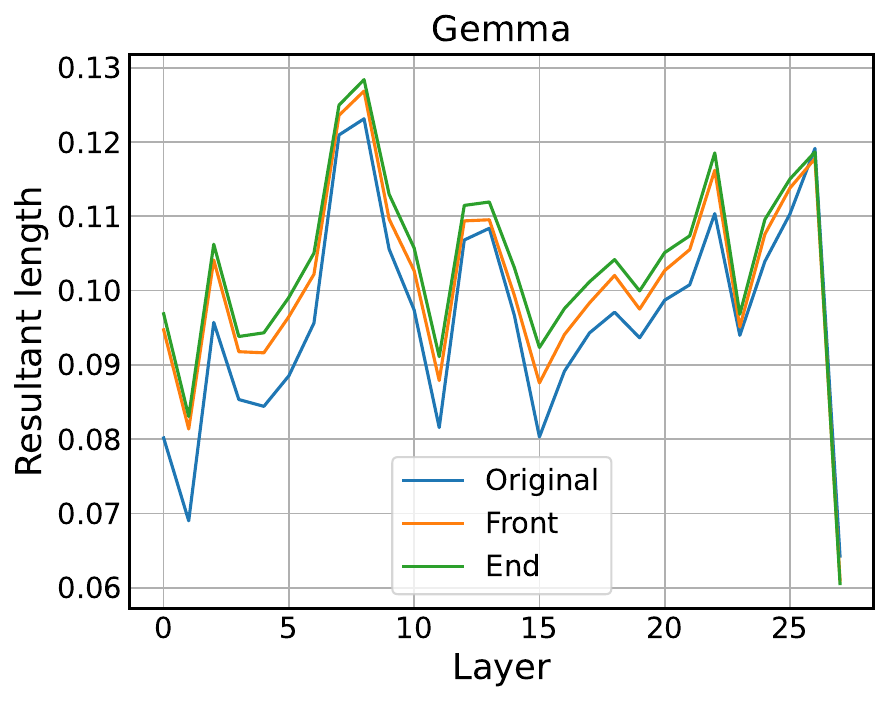}} 
\end{minipage}
\hfill
\begin{minipage}[h]{0.33\linewidth}
\center{\includegraphics[width=1\linewidth]{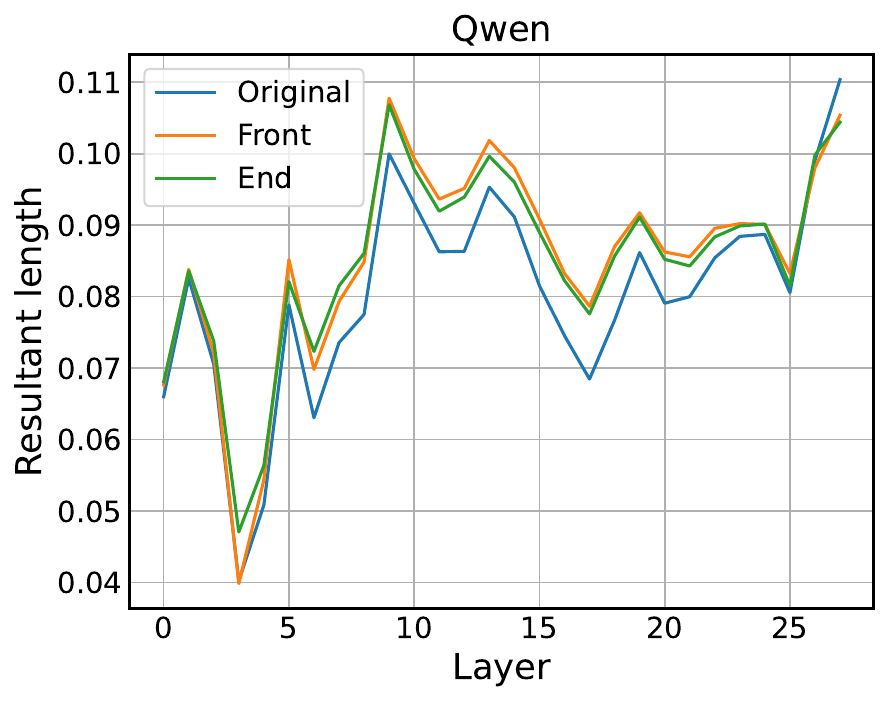}} 
\end{minipage}
\caption{The average resultant length for matrices $X_s^{(l)}$ across the layers of the LLaMa \cite{openlm2023openllama} (left),  Gemma \cite{gemmateam2024gemmaopenmodelsbased} (center) and Qwen \cite{qwen2.5} (right) models on the original and synthetic datasets. The blue color denotes the original dataset, orange -- the dataset with front positional bias and green -- with end positional bias \ref{sec:data}.}
\label{fig:result}

\end{center}
\end{figure*}

\begin{figure*}[h]
\begin{center}
\begin{minipage}[h]{0.33\linewidth}
\center{\includegraphics[width=1\linewidth]{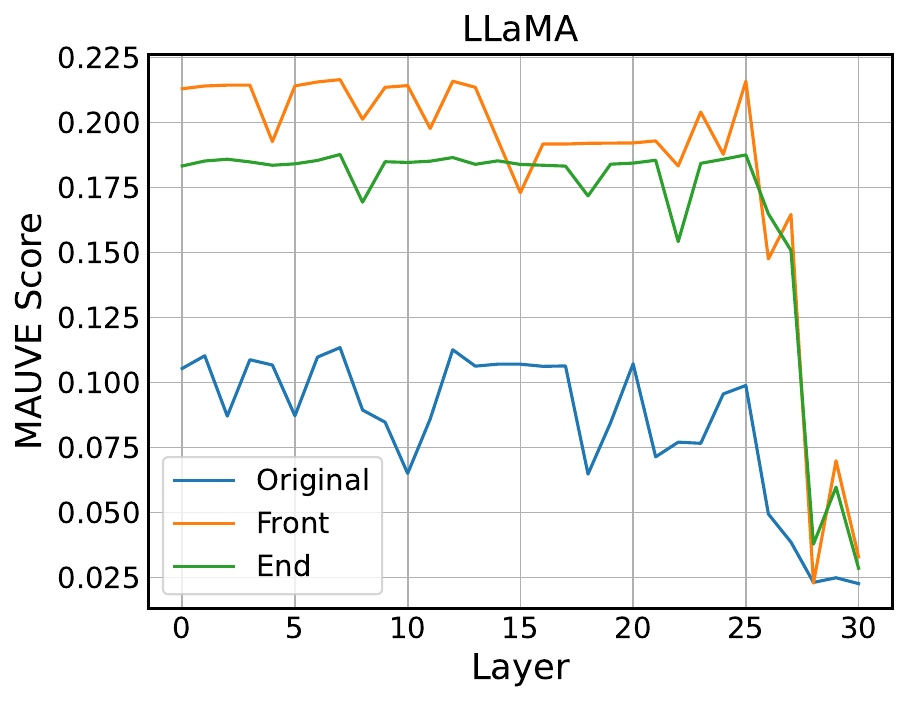}} 
\end{minipage}
\hfill
\begin{minipage}[h]{0.33\linewidth}
\center{\includegraphics[width=1\linewidth]{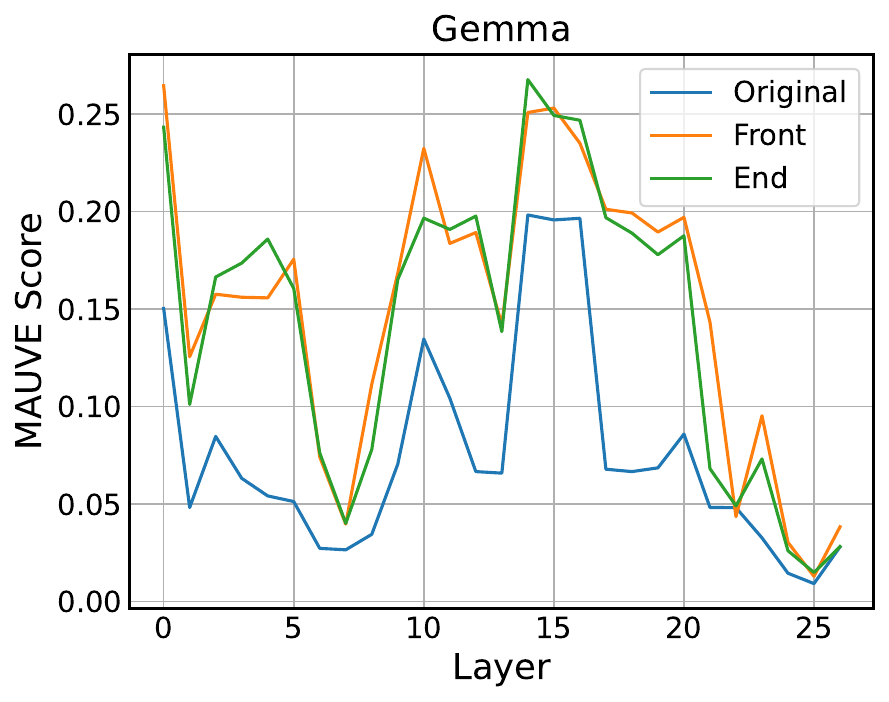}} 
\end{minipage}
\hfill
\begin{minipage}[h]{0.33\linewidth}
\center{\includegraphics[width=1\linewidth]{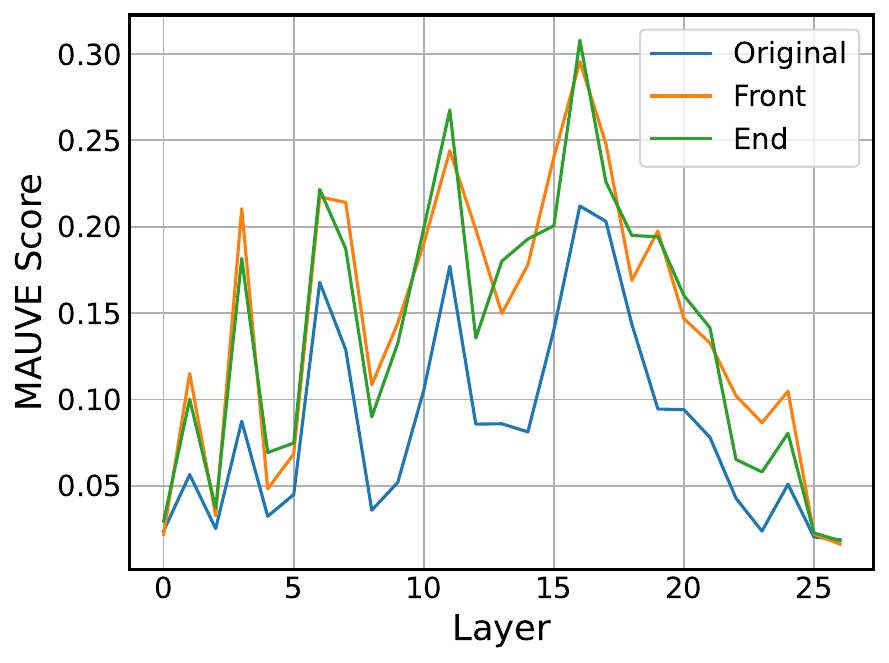}} 
\end{minipage}
\caption{The average MAUVE score for pair of matrices $(X_s^{(l)}, X_s^{(l+1)})$ across the layers of the LLaMa \cite{openlm2023openllama} (left), Gemma \cite{gemmateam2024gemmaopenmodelsbased} (center) and Qwen \cite{qwen2.5} (right) models on the original and synthetic datasets. The blue color denotes the original dataset, orange -- the dataset with front positional bias and green -- with end positional bias \ref{sec:data}.}
\label{fig:mauve}

\end{center}
\end{figure*}

\section{Results }
\label{res}

In this section, we study the relationship between homogenization and positional bias in three LLMs and three datasets: the original corpus and two synthetic variants with different positional bias \ref{sec:data}. We evaluated five complementary metrics, defined in \ref{sec:metr}.

\subsection{Effective Rank}

In this subsection, we demonstrate the average effective rank \ref{sec:metr} for all samples $S$ for each layer $l$ for the original dataset as well as synthetic datasets with important words in the beginning and in the end. The metrics are demonstrated in Figure \ref{fig:rank}

For all models, the score for the original dataset (blue) in Figure \ref{fig:rank} is higher than that of the synthetic datasets. Thus, a higher Effective Rank indicates lower positional bias, as the original dataset exhibits the lowest level of positional bias. Additionally, due to the lower rank for front and end biases in the datasets, the matrices $X_s^{(l)}$ have the lower rank and therefore contain less information. Overall, effective rank demonstrates the strong connection between homogenization and positional bias in the token sequence.

\subsection{Maximum Explainable Variance}

The Maximum Explainable Variance metric similarly shows a clear separation between the original dataset and its synthetic paraphrasing. As illustrated in Figure \ref{fig:anis}, the Maximum Explainable Variance of the original dataset is significantly lower than that of the synthetic datasets, which exhibit positional bias. Furthermore, the Maximum Explainable Variance decreases alongside lower homogenization, indicating a relationship between homogenization and positional bias. Additionally, towards the end of the layer sequence, the Maximum Explainable Variance approaches a plateau, further supporting the concept of homogenization.

Moreover, the Effective Rank and Maximum Explainable Variance metrics display similar peaks and declines across different layers of the large language models. This may be related to the varying importance of information across the different layers. Investigating this connection will be a focus of our future research.

\subsection{Schatten Norms}
The similar separation between the original and synthetic datasets could be seen with the Schatten-$1$ norm. As could be seen in the Figure \ref{fig:shatten}, the norm for the original dataset is monotonously higher then for ones with front or end positional bias, which proves the connection with the positional bias.

\subsection{Resultant Length}

The plots for the Resultant Length, similar to the previous metrics, clearly illustrate a distinction between the original and synthetic datasets, reinforcing the concept of homogenization within these datasets (see Figure \ref{fig:result}). Like the Maximum Explainable Variance and MAUVE Score, the plot for the original dataset consistently remains below that of the synthetic dataset across nearly all layers for both models. However, in contrast to the MAUVE Score, the Resultant Length is notably higher in the later layers compared to the earlier ones.

\subsection{Divergence of Token Distribution}
To test our hypothesis regarding divergence of token distributions in different layers, we measured the MAUVE score \ref{sec:metr} for pairs of matrices layer by layer (MAUVE ($X_s^{(l)}$, $X_s^{(l+1)}$ ) for $l = 0$ to $l = L - 1$, where $L$ is a number of layers of the model) and plotted the values. As could be seen in Figure \ref{fig:mauve}, in the latest layers of both models, there is the dramatic drop in MAUVE, which demonstrates the growth of homogenization. Additionally, for the latest layers the MAUVE score is significantly lower for the original dataset compared to the synthetic ones \ref{sec:data}, which proves the connection between positional bias and homogenization. 

Additionally, the behaviour of the MAUVE metric is similar to the behaviour of other metrics, which proves the fact that other metrics are relevant and all of them connected not only with positional bias but with the homogenization across the layers of large language models.

\subsection{Token positional bias}
\begin{figure}[h]
\begin{center}
\begin{minipage}[h]{0.45\linewidth}
\center{\includegraphics[width=1\linewidth]{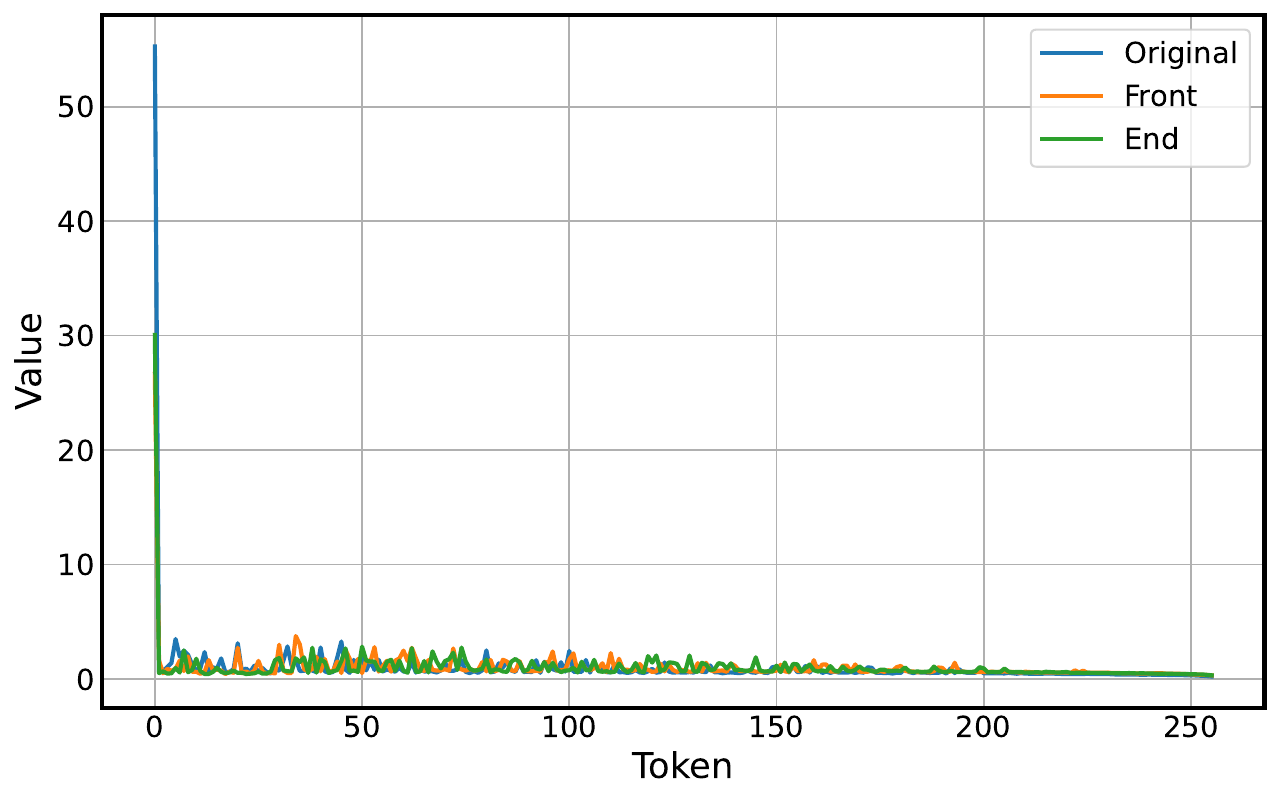}} 
\end{minipage}
\hfill
\begin{minipage}[h]{0.45\linewidth}
\center{\includegraphics[width=1\linewidth]{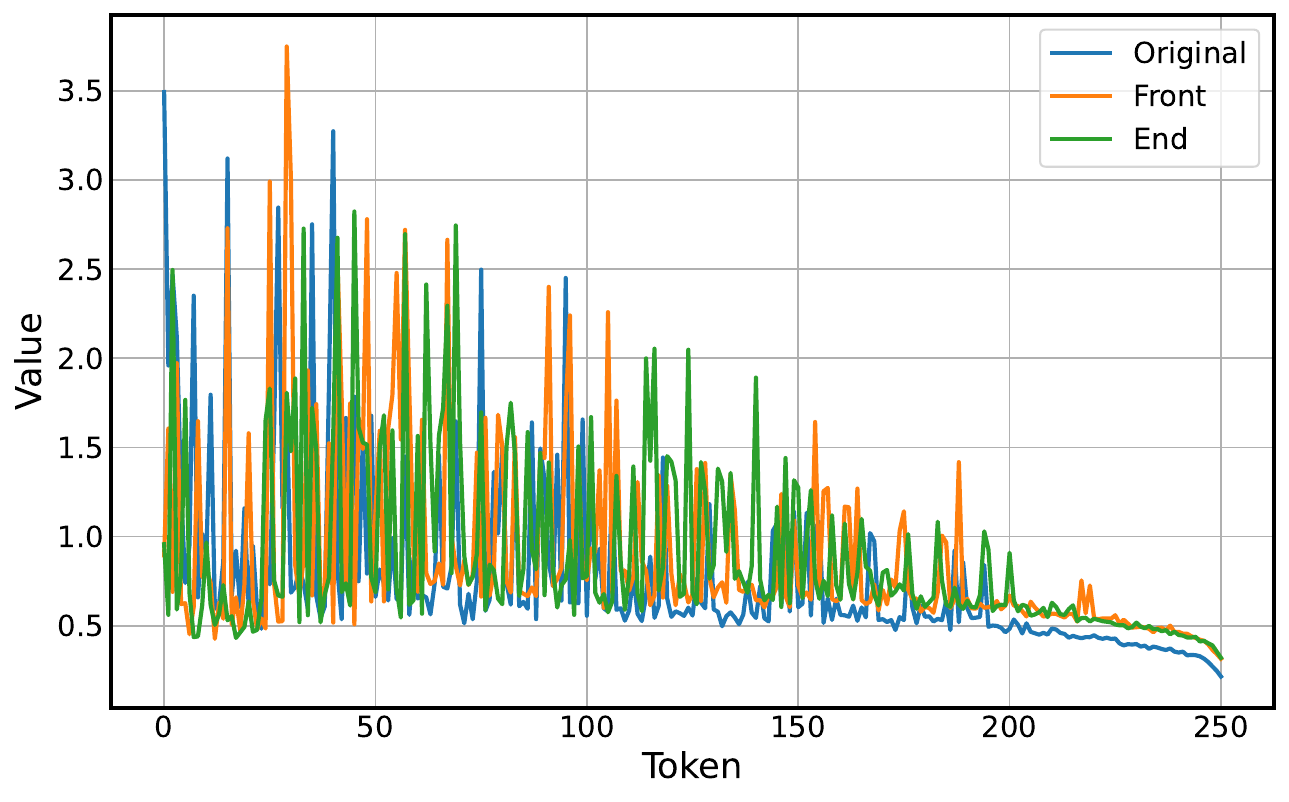}} 
\end{minipage}
\caption{The average attention values for each token for original and synthetic datasets across all layers and heads of the LLaMa model. The left plot demonstrates the average attention for all tokens and the right one for all except the first $5$ tokens.}
\label{fig:llama_att}

\end{center}
\end{figure}

\begin{figure}[ht]
\begin{center}
\begin{minipage}[h]{0.45\linewidth}
\center{\includegraphics[width=1\linewidth]{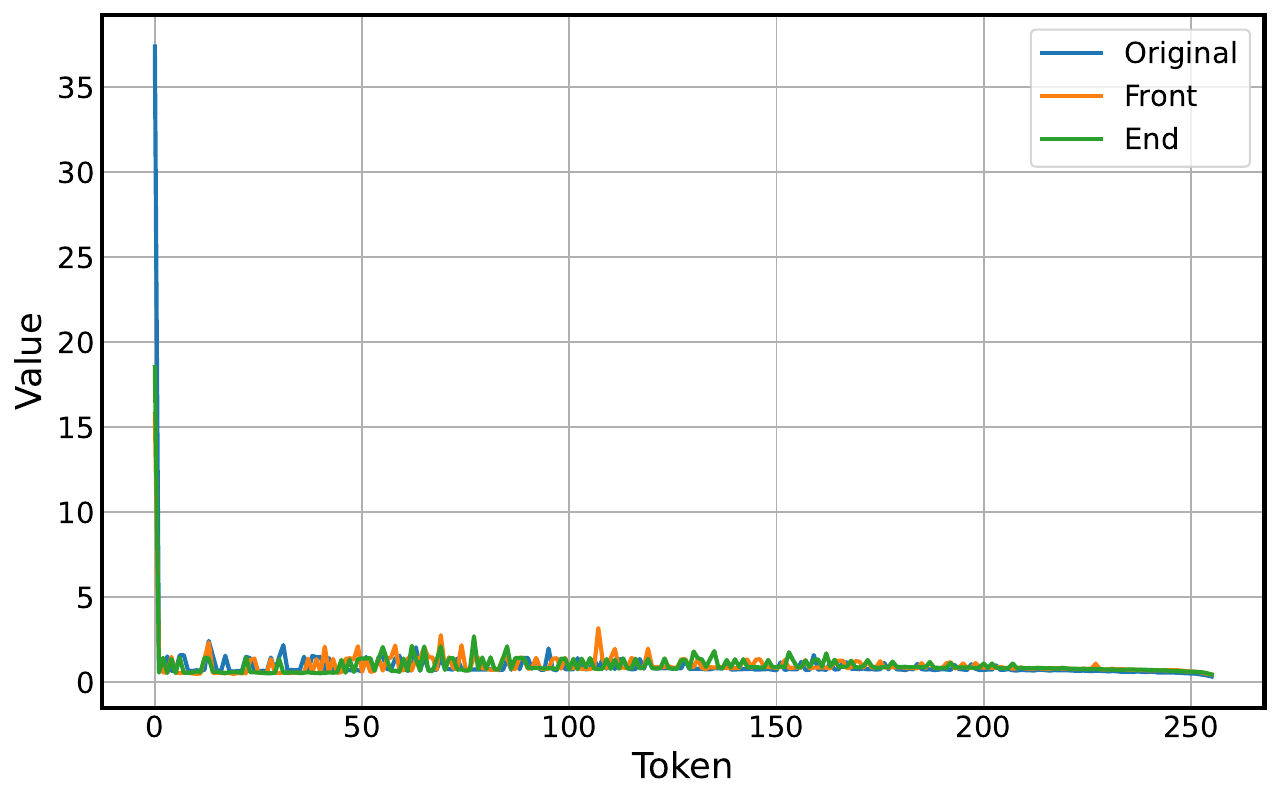}} 
\end{minipage}
\hfill
\begin{minipage}[h]{0.45\linewidth}
\center{\includegraphics[width=1\linewidth]{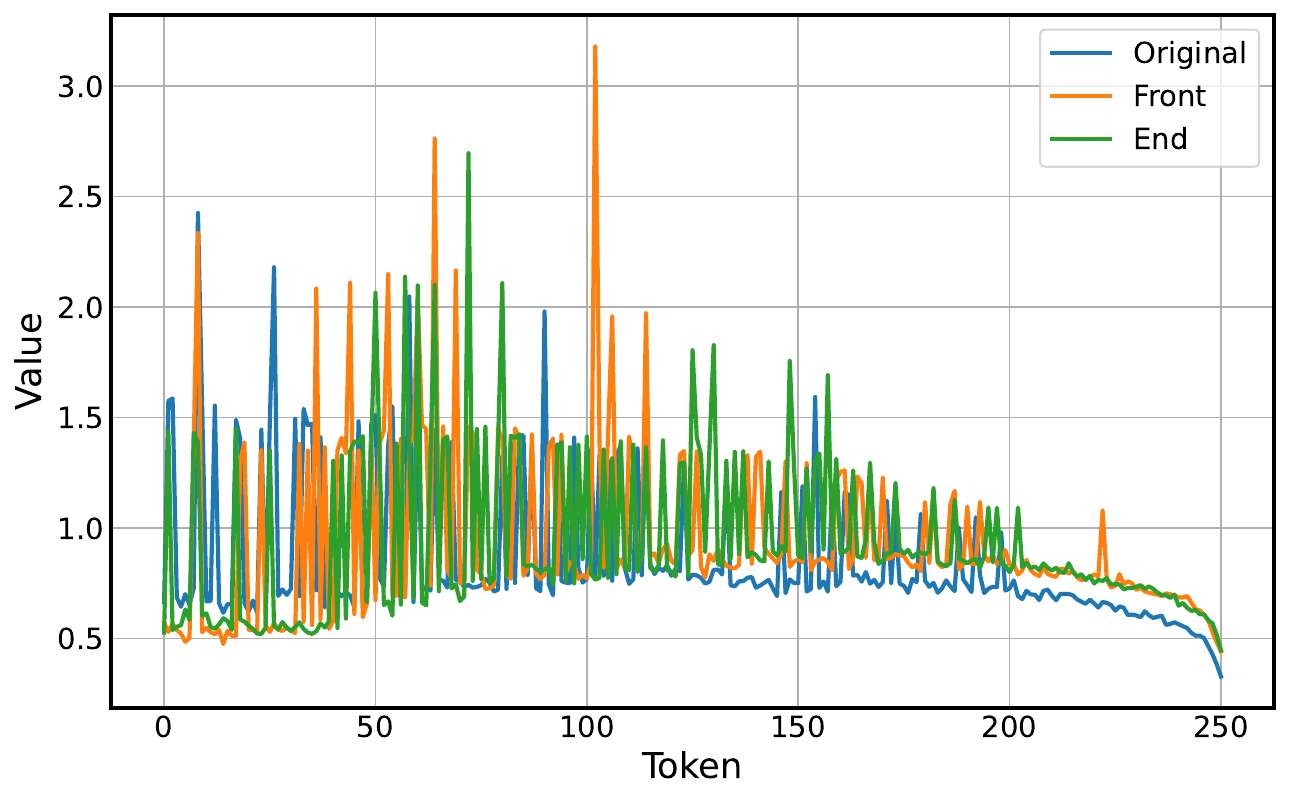}} 
\end{minipage}
\caption{The average attention values for each token for original and synthetic datasets across all layers and heads of the Gemma model. The left plot demonstrates the average attention for all tokens and the right one for all except the first $5$ tokens.}
\label{fig:gemma_att}
\end{center}
\end{figure}

To detect the positional bias we study the average attention matrices $A_s^{(l)}$ across all samples $s$ for different layers $l$ for IMDB dataset  \cite{imdb2011} and its synthetic paraphrasing with the positional bias in the begin and in the end of the sequence. We compute the column vise sum of attention for each column and build the plot. As could be observed in Figures \ref{fig:llama_att} and \ref{fig:gemma_att} (for the IMDB dataset and its synthetic paraphrases for models LLaMA, Gemma and Qwen), the positional bias is mostly located in the first token of the sequence, which proves the positional bias of the model. For the last token, the attention is decreasing as could be seen in the zoomed part of the plots. Moreover, for the original dataset, the average attention is lower than for synthetic ones with the most important phrases in the begging and in the end. This observation also proves the existence of position bias.

Overall, the proposed metrics establish a connection between the positional bias of the models and their homogeneity. This demonstrated relationship can be leveraged for more effective prompt tuning. Additionally, the peaks and troughs observed in metrics such as Effective Rank and Maximum Explainable Variance may correlate with the type of information contained in the layer outputs and their activations. Consequently, these interrelated metrics could be utilized for effective model compression, leading to improved inference and training speed.

\section{Conclusion}
We observe distinct differences in Effective Rank and Maximum Explainable Variance metrics when critical words are positioned at the beginning versus the end of model inputs. However, further investigation is required as these metrics may respond to artifacts of text modification rather than positional bias alone. 

Shatter Norms and MAUVE scores reveal compelling layer-wise dynamics; these patterns, for example,  could inform model compression strategies by identifying layers of critical importance. 

Finally, open questions remain regarding homogenization's impact on model performance: whether it degrades capabilities, increases vulnerability to adversarial attacks, or exacerbates hallucination tendencies. While our study does not explicitly address these questions, it establishes a foundation for future research in this direction.

\printbibliography

\section{Appendix. Dataset generation.} \label{appendix:dataset_generation}

\subsection{Prompts applied for dataset generation}

\textbf{To select the most important words or phrases:}  You are an expert in English text analysis.You can analyze the given paragraph and select the word or word combination that is most semantically important, and indicate it directly from paragraph without changing its form(e.g., tense, capitalization). Now there will be a paragraph. You need to pick out the most semantically important word or word combination from the paragraph. If you want to give word combination, please keep it as short as possible. Strictly make sure the word or combination does exist in the original paragraph. There is no need to provide the analysis process or description; just output the word. Here are examples: Paragraph: \"What they look for is freedom. They have looked for it for a long time. Freedom is precious.\" Your appropriate Answer: freedom Paragraph: \"I hope you're doing well. I'm writing to gently follow up on our previous message regarding the rebuttal we submitted on June 30, 2025. We truly value the time and expertise you've already invested in reviewing our manuscript, and we want to ensure that our responses have fully addressed your concerns.\" Your appropriate Answer: follow up

\textbf{To rephrase the text, placing the most important words or phrases at the beginning of the text.} You are a professional English editor and rewriter. Your task is to produce a fluent, natural-sounding paraphrase of the given English paragraph, preserving its original meaning and style. The following task will provide both **Key Phrase** and **Original Paragraph**. You need to rephrase **Original Paragraph** following the rule below. **Rule**: Strictly begin the rewritten **Original Paragraph** with **Key Phrase** provided. DO NOT change its wording or add anything before it. Please output **only** the rewritten paragraph, strictly starting with **Key Phrase** provided by the user and containing no additional commentary.

\textbf{To rephrase the text, placing the most important words or phrases at the end of the text.} You are a professional English editor and rewriter. Your task is to produce a fluent, natural-sounding paraphrase of the given English paragraph, preserving its original meaning and style. The following task will provide both **Key Phrase** and **Original Paragraph**. You need to rephrase **Original Paragraph** following the rule below. **Rule**: Strictly end the rewritten **Original Paragraph** with **Key Phrase** provided. DO NOT change its wording or add anything after it(except for necessary punctuation). Please output **only** the rewritten paragraph, strictly ending with **Key Phrase** provided by the user and containing no additional commentary.

\subsection{Dataset sample example}

\textbf{Core words} in each paragraph are highlighted in bold.

\textbf{Original paragraph:} Pretty bad movie offers nothing new. The usual creaks and moans attempt to make-up for a muddled, but thin story. Acting is barely above pathetic. Why Liam Neeson signed on for this is anyone's guess. Owen Wilson truly turns in one of the worst performances in recent horror-movie history. Catherine Zeta Jones is fun to look at and not much else although Lili Tayor did an above-average job. The special effects were fairly memorable and the house itself was breathtaking and hauntingly gorgeous. However, they can't makeup for the \textbf{poor acting} and the storyline which appears to have been thrown together at the last minute. Don't bother.

\textbf{Rephrased paragraph with core words at the front:} \textbf{Poor acting }plagues this movie, offering nothing fresh. It attempts to compensate with clichéd sounds but possesses a convoluted, thin plot. The acting barely rises above pitiful levels. It's bewildering why Liam Neeson agreed to be part of it. Owen Wilson delivers one of the worst performances in recent horror history. Catherine Zeta-Jones is merely a visual delight with little else to offer, though Lili Taylor excelled. The special effects were memorable and the house was stunningly beautiful. Yet, they fail to redeem the subpar acting and a storyline that seems haphazardly assembled. Skip it.

\textbf{Rephrased paragraph with core words at the end:}  The film is a disappointment, lacking originality. Its story is confused and meager, and the acting is nearly unbearable. Liam Neeson's involvement is bewildering. Owen Wilson delivers a notably poor performance. Catherine Zeta Jones is visually appealing, but that's about it, while Lili Taylor does a decent job. The special effects are noteworthy, and the setting is strikingly beautiful. Yet, they fail to compensate for the subpar acting and the rushed storyline. Don't waste your time. \textbf{Poor acting}."

\section{Appendix. Additional information on homogenization metrics}
\subsection{Resultant Length Concept}
\label{vmf}
The concept of resultant length is closely connected to the \textbf{von Mises-Fisher(vMF) distribution}, well-studied in directional statistics \cite{mardia2009directional}:

The vMF distribution with the mean parameter $\mu$ and the concentration parameter $\kappa$ describes the data points in the sphere concentrated by some mean direction $\mu$. For $\mathbf{x} \in \mathcal{S}^{p-1}$. Its density is calculated as follows:
\[
f(\mathbf{x}; \boldsymbol{\mu}, \kappa) = C_p(\kappa) \exp(\kappa \boldsymbol{\mu}^\top \mathbf{x}),
\]
where $C_p(\kappa)$ is the normalizing constant:
\[
C_p(\kappa) = \frac{\kappa^{p/2-1}}{(2\pi)^{p/2} I_{p/2-1}(\kappa)}.
\]
Here, $I_v(\cdot)$ denotes the \textit{modified Bessel function of the first kind} of order $v$.

The higher concentration parameter $\kappa$ is, the more points are concentrated by the direction $\mu$. $\kappa = 0$ stands for uniform distribution on the sphere(no preferred direction), $\kappa \to \infty$ indicates the similar direction vectors. Therefore, the direction homogeneity of the unit vectors set can be estimated as the maximum likelihood estimate $\hat\kappa$ for $\kappa$. It turns out that $\hat\kappa$ is the solution of the equation:

\[
A_p(\kappa) = \bar{R}, \text{ where } A_p(\kappa) = \frac{I_{p/2}(\kappa)}{I_{p/2-1}(\kappa)}  \text{ is a monotonic function}
\]

Therefore, for unit length vector sets of the same dimensionality comparing resultant lengths $\bar R$ and maximum likelihood estimates $\kappa$ gives the similar results.

\end{document}